\documentclass[10pt,twocolumn,letterpaper]{article}

\usepackage{iccv}              

\definecolor{iccvblue}{rgb}{0.21,0.49,0.74}
\usepackage[pagebackref,breaklinks,colorlinks,allcolors=iccvblue]{hyperref}

\def\paperID{*****} 
\def\confName{ICCV}
\def\confYear{2025}

\title{Kaleidoscopic Background Attack: Disrupting Pose Estimation \\ with Multi-Fold Radial Symmetry Textures}

\author{
    Xinlong Ding$^{*1}$, Hongwei Yu$^{*1}$, Jiawei Li$^{*1}$, Feifan Li$^{1}$, Yu Shang$^{2}$\\ 
    Bochao Zou$^{1}$, Huimin Ma$^{1}$, Jiansheng Chen$^{\dagger1}$\\
    \small{$^1$ University of Science and Technology Beijing, China \quad 
        $^2$ Tsinghua University, China} \\
    \definecolor{github}{RGB}{142, 36, 170}
    \hypersetup{urlcolor=github}
    \small{\textbf{\url{https://wakuwu.github.io/KBA}}}
}

\begin{document}
\maketitle

\begin{abstract}
Camera pose estimation is a fundamental computer vision task that is essential for applications like visual localization and multi-view stereo reconstruction.
In the object-centric scenarios with sparse inputs, the accuracy of pose estimation  can be significantly influenced by background textures that occupy major portions of the images across different viewpoints.
In light of this, we introduce the Kaleidoscopic Background Attack (KBA), which uses identical segments to form discs with multi-fold radial symmetry. 
These discs maintain high similarity across different viewpoints, enabling effective attacks on pose estimation models even with natural texture segments.
Additionally,  a projected orientation consistency loss is proposed to optimize the kaleidoscopic segments, leading to significant enhancement in the attack effectiveness. 
Experimental results show that optimized adversarial kaleidoscopic backgrounds can effectively attack various camera pose estimation models.
\end{abstract}
{
  \renewcommand{\thefootnote}{\fnsymbol{footnote}}
  \footnotetext[0]{* Equal contribution.\quad $\dagger$ Corresponding author (jschen@ustb.edu.cn).} 
} 
\section{Introduction}
\label{sec:intro}

Camera pose estimation involves determining the positions and orientations of cameras based on multi-view images. The accuracy of these estimates is critical for various downstream tasks, including visual localization~\cite{Giang2024vl1, Jiang2024vl2}, multi-view stereo reconstruction~\cite{Das2024sr1, Xie2024sr2}, and novel view synthesis~\cite{ Lu2024nvs1, Lee2024nvs2}. 
Sparse-view object-centric scenes, where objects are centered on a flat surface and imaged by cameras oriented towards them, are among the most common scenarios in practical applications.
Classic methods like Structure from Motion (SfM)~\cite{schaffalitzky2002multi_sfm1, Agarwal2009_sfm2, Crandall2011_sfm3, Wilson2014_sfm5} can not adapt to such scenarios since they require dense viewpoints.
Consequently, many learning-based approaches~\cite{Wang2024dust3r, vincent2024mast3r, zhang2024raydiffusion, Wang2023PoseDiffusion, lin2024relposepp, Sinha2023CVPR_lbpe1, Zhang2022ECCV_lbpe2, Hanwen2022_lbpe3, Rockwell2022_lbpe5} have been proposed to bridge this gap, achieving satisfying performance with sparse views.

However, these learning-based methods often rely on background information that occupies a large portion of the image to accurately estimate camera poses from sparse views.
This reliance makes it possible for specific texture patterns to interfere with the model's output. However, the robustness of pose estimation models under such situations has not been fully discussed previously.
In light of this, we aim to explore the vulnerability of such models by leveraging background textures through adversarial attacks.

\begin{figure}[t]
  \centering
   \includegraphics[width=0.949\linewidth]{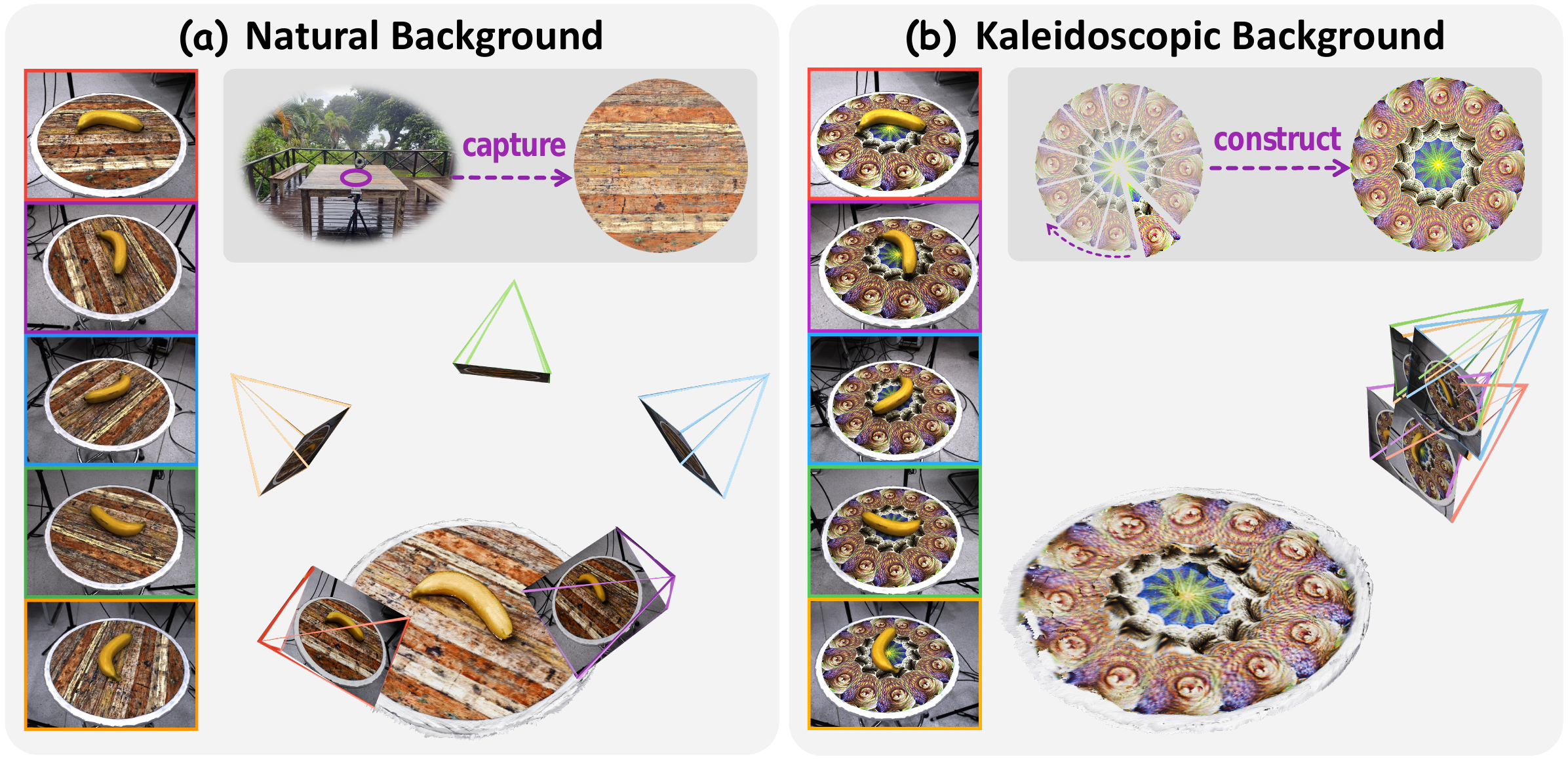}
   \caption{
Impact of natural and kaleidoscopic backgrounds on camera pose estimation in object-centric scenes. (a) With a natural tabletop background, the DUSt3R model accurately estimates the camera pose and reconstructs the banana. (b) With a kaleidoscopic background, the model predicts erroneous but similar poses across viewpoints, leading to reconstruction failure.
   }
   \label{fig:first}
\end{figure}

Since the pioneering study by Szegedy et al.~\cite{AdvPioneer}, the adversarial robustness of deep neural networks (DNNs) has been extensively studied over the years~\cite{FGSM, IFGSM, PGD, C&W, MI, ding2024transferable, yu2024step, li2025A2RNet}. 
Classic patch-based attacks~\cite{classic_patch1,classic_patch2,rw_pa_patch,rw_pa_patch3,rw_pa_patch4,ding2024Invisible}, a common form of adversarial attack, often optimize perturbations directly on a standalone patch image, limiting their adaptability in physical environments. 
Recent studies try to address these limitations by leveraging repeated texture patterns~\cite{hu22cloth,rw_pa_patch9} and learnable patch shapes and locations~\cite{rw_pa_patch2,wei23cloth}, significantly enhancing their success rates in the physical world.
This leads us to consider what kinds of texture priors can enhance adversarial attacks for camera pose estimation tasks in the physical world.

We observe that numerous radial symmetric patterns exist in nature and everyday life~\cite{ball2009natures}, such as five-fold starfish, six-fold snowflakes, and various multi-fold patterns like flowers, water splashes, and kaleidoscopes.
Such radially symmetric textures enable background similarity across multiple viewpoints.
Inspired by this, we select a radially symmetric disc, uniformly divided into several segments, resembling a sliced pizza. 
Each of the \( N \) segments shares the same texture, forming an \( N \)-fold radially symmetric kaleidoscopic disc that provides consistent background appearances across multiple viewpoints, as illustrated in Fig.~\ref{fig:first}(b).
We begin with natural textures scanned from a desktop as segments to create the kaleidoscopic disc, referred to as KBA\textsubscript{nat} in the following sections.
Experimental results show that a kaleidoscopic disc constructed solely from natural texture segments can already noticeably impact camera pose estimation across various models.

To further enhance the adversarial impact of the kaleidoscopic background on camera pose estimation, we introduce adversarial attacks to craft the kaleidoscopic segments, resulting in an optimized radially symmetric background, referred to as KBA\textsubscript{opt}.
Specifically, we leverage differentiable rendering techniques~\cite{ravi2020pytorch3d} to generate diverse multi-view object-centric scenes, incorporating various objects and background environments to simulate real-world scenarios.
Building on this, we select a popular camera pose estimation model as a surrogate to conduct adversarial attacks by maximizing camera orientation similarity across different viewpoints.
Experiments demonstrate that the optimized segments exhibit improved radial symmetry when forming a kaleidoscopic disc, leading to significantly greater effectiveness and stability in disrupting various camera pose estimation models compared to the non-optimized KBA\textsubscript{nat}.
\noindent 
In summary, our contributions are as follows:
\begin{itemize}
\item 
Inspired by the prevalent symmetry in nature, we propose a method to construct adversarial kaleidoscopic background with multi-fold radial symmetry in object-centric scenes to effectively disrupt camera pose estimation.
\item 
We optimize the kaleidoscopic background using orientation consistency loss to significantly enhance the attack effectiveness in both the digital and physical worlds.
\item 
To the best of our knowledge, we are the first to utilize background textures as adversarial examples to attack sparse-view camera pose estimation models. 
Our work introduces a method for constructing challenging samples, which can facilitate improvement in both the performance and robustness of these models in the future.
\end{itemize}

\begin{figure*}[ht]
  \centering
   \includegraphics[width=\linewidth]{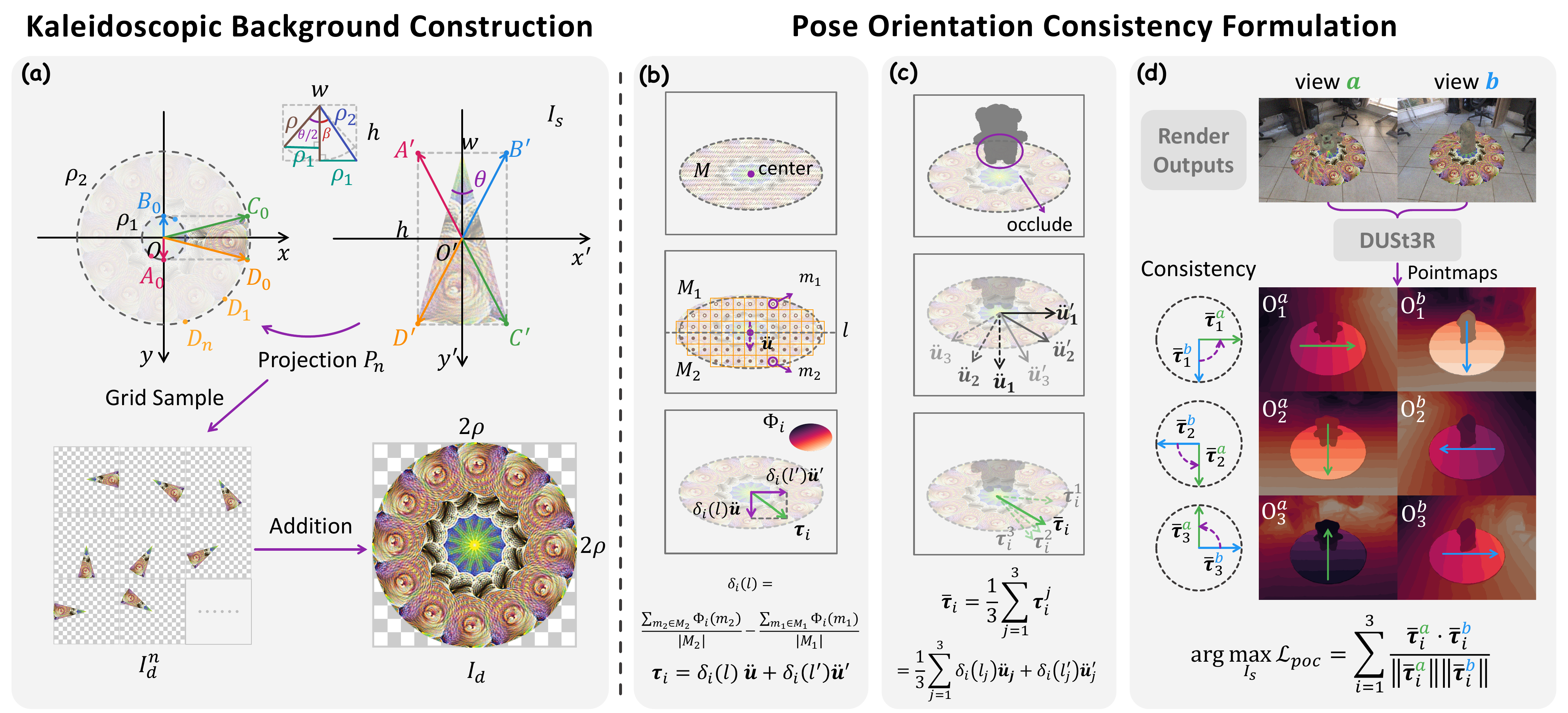}
\caption{(a) Construction of the kaleidoscopic background disc. (b) Estimation of coordinate flow direction. (c) Calculation of average flow directions across multiple bisections. (d) Computation of the projected orientation consistency loss for camera pose estimation.}
   \label{fig:method}
\end{figure*}

\section{Method}
\label{sec:method}
In this section, we first introduce the construction of the multi-fold symmetric kaleidoscopic background, which serves as the foundation of our approach. 
Leveraging this construction, using natural textures as segments can also generate considerable interference effects on various models. 
To enhance the effectiveness of perturbations, we select DUSt3R~\cite{Wang2024dust3r}, a model capable of performing various 3D tasks that has garnered widespread attention from the research community, as our target for white-box adversarial attacks.
We will introduce how to use the proposed projected orientation consistency loss to constrain the camera orientations from any two viewpoints, thereby optimizing the segments to construct a radially symmetric background.

\subsection{Kaleidoscopic Background Construction}
\label{sec:pattern}
Unlike conventional adversarial attacks that use a single image as the adversarial example, we construct an adversarial background disc \( I_d \in \mathbb{R}^{2\rho \times 2\rho \times 3} \) with a radius \(\rho\) from \(N\) segment images.
To achieve this, we begin by initializing a segment image \( I_s \in \mathbb{R}^{w \times h \times 3} \), as shown in Fig.~\ref{fig:method}(a), with height \( h \), width \( w \), and segment angle \( \theta \) computed using Eq.~\ref{eq:hw}, where \( \lceil \cdot \rceil \) denotes the ceiling function.
\begin{equation}
\theta = \frac{2\pi}{N}, \quad
h = \rho, \quad
w = \lceil 2 \rho \sin{\frac{\theta}{2}} \rceil
\label{eq:hw}
\end{equation}
For an arbitrary segment of \( I_d \), the corresponding region can be obtained by projecting the image \( I_s \) via perspective projection.
In this process, solving the perspective projection matrix for mapping the \( x'O'y' \) coordinate system to \( xOy \) reduces to determining the vertices of the rectangular regions \( A'B'C'D' \) and \( A_nB_nC_nD_n \) with the help of the OpenCV~\cite{opencv_library} library.
Given the height \( h \) and width \( w \) of \( I_s \), \( \{A',B',C',D'\}\) can be easily calculated as shown in Eq.~\ref{eq:seg_abcd}.
\begin{equation}
\{(-\frac{h}{2}, -\frac{w}{2}),\, (\frac{h}{2}, -\frac{w}{2}),\, (\frac{h}{2}, \frac{w}{2}),\, (-\frac{h}{2}, \frac{w}{2})\}
\label{eq:seg_abcd}
\end{equation}
As for the \( n \)-th rectangle \( A_nB_nC_nD_n \), it can be derived by rotating \( A_0B_0C_0D_0 \) around the origin \( O \) by an angle of \( n\theta \). 
Consequently, vertices \( A_n \) and \( B_n \) lie on a circle with radius \( \rho_1 \), while \( C_n \) and \( D_n \) lie on a circle with radius \( \rho_2 \), where \( \rho_1 \) and \( \rho_2 \) are given by Eq.~\ref{eq:r1r2}. 
\begin{equation}
\rho_1 = \rho\sin{\frac{\theta}{2}}, \quad 
\rho_2 = \sqrt{\rho_1^2 + \rho^2}
\label{eq:r1r2}
\end{equation}
The coordinates of vertices \( A_nB_nC_nD_n \) can further be expressed using Eq.~\ref{eq:disc_abcd}, where \( \beta = \arctan(\rho_1 / \rho) \).
\begin{equation}
\begin{cases} 
A_n = \left(\rho_1\cos{(n\theta+\frac{\pi}{2})},\ \rho_1\sin{(n\theta+\frac{\pi}{2})}\right)  \\
B_n = \left(\rho_1\cos{(n\theta-\frac{\pi}{2})},\ \rho_1\sin{(n\theta-\frac{\pi}{2})}\right)  \\
C_n = \left(\rho_2\cos{(n\theta-\beta)},\ \rho_2\sin{(n\theta-\beta)}\right)  \\
D_n = \left(\rho_2\cos{(n\theta+\beta)},\ \rho_2\sin{(n\theta+\beta)}\right)
\end{cases}
\label{eq:disc_abcd}
\end{equation}
By combining Eq.~\ref{eq:seg_abcd} and Eq.~\ref{eq:disc_abcd}, the transformation matrix \( P_n \) that maps any point in \( A'B'C'D' \) to \( A_nB_nC_nD_n \) can be easily obtained by calling the \texttt{getPerspectiveTransform} function in OpenCV.
The projected image \( I_d^n \in \mathbb{R}^{2\rho \times 2\rho \times 3} \), obtained by applying \( P_n \) to the segment image \( I_s \), can be derived from Eq.~\ref{eq:seg2disci}, where \( \mathcal{G}(I_d) \) represents the grid of the image \( I_d \), and each element in \( P_n^{-1} \mathcal{G}(I_d) \in \mathbb{R}^{2\rho \times 2\rho \times 3} \) denotes a sampling coordinate from \( I_s \). 
\( G(\cdot) \) can be implemented using the \texttt{grid\_sample} function in the PyTorch~\cite{pytorch2} library.
\begin{equation}
I_d^n = G(I_s, P_n^{-1} \mathcal{G}(I_d))
\label{eq:seg2disci}
\end{equation}
Finally, the kaleidoscopic background image \( I_d \) can be generated by performing element-wise addition on the \( N \) projected images, as shown in Eq.~\ref{eq:seg2disc}.
\begin{equation}
I_d = \sum_{n=0}^{N-1} I_d^n
\label{eq:seg2disc}
\end{equation}
\textbf{In fact, a kaleidoscopic background  \( I_d \) composed of segments of natural texture is already capable of attacking the pose estimation models. }
However, further optimizing \( I_d \)  can enhance the attack effectiveness.
In the following section, we will introduce the design and physical meaning of the proposed optimization loss.

\subsection{Enforced Orientation Consistency Loss}
\label{sec:loss}
\noindent
\textbf{Ideal enforced orientation consistency loss.}
The camera pose is typically represented by a rotation matrix \( R \in \mathbb{R}^{3 \times 3} \) for the camera orientation and a translation vector \( T \in \mathbb{R}^{3 \times 1} \) for the position, as shown in Eq.~\ref{eq:R_matrix} and Eq.~\ref{eq:T_vector}. 
Vectors \( \mathbf{r}_1 \), \( \mathbf{r}_2 \), and \( \mathbf{r}_3 \) in \( \mathbb{R}^{3 \times 1} \), formed by the rows of \( R \), represent the directions of the camera coordinate system axes.
\begin{equation}
R =
\begin{bmatrix} 
r_{11} & r_{21} & r_{31} \\
r_{12} & r_{22} & r_{32} \\
r_{13} & r_{23} & r_{33}
\end{bmatrix}^\top =
\begin{bmatrix} \mathbf{r}_1 \hspace{-5pt} & \mathbf{r}_2 \hspace{-5pt} & \mathbf{r}_3  \end{bmatrix}^\top
\label{eq:R_matrix}
\end{equation}
\begin{equation}
T = \begin{bmatrix} t_1 \hspace{-5pt} & t_2 \hspace{-5pt} & t_3 \end{bmatrix}^\top
\label{eq:T_vector}
\end{equation}
For pose estimation models that directly output the matrix \( R \), an ideal attack can be achieved by maximizing the sum of cosine similarities between the corresponding vectors \( \mathbf{r}_i \) from two views \( a \) and \( b \), as detailed in Eq.~\ref{eq:ideal_ocl}.
Here, \(\|\cdot\|\) denotes the \( L_2 \) norm of a vector. 
\textbf{Such an attack aims to enforce the convergence of pose orientations from different views to a single direction.}
Thereby, a multi-view imaging system degrades into a single-view system in a sense.
Such degradation will significantly impact downstream tasks by disrupting the restoration of spatial information.
\begin{equation}
\mathcal{L}_{oc} = \sum_{i=1}^3 \frac{\mathbf{r}_i^a \cdot \mathbf{r}_i^b}{\|\mathbf{r}_i^a\| \|\mathbf{r}_i^b\|}
\label{eq:ideal_ocl}
\end{equation}
However, recent models such as DUSt3R~\cite{Wang2024dust3r} and MASt3R~\cite{vincent2024mast3r} output pointmaps instead of \( R \) matrix, leading to the difficulty in directly applying Eq.~\ref{eq:ideal_ocl}.
We define a point in the world, camera, and pixel coordinate systems as \((x, y, z)\), \((\dot{x}, \dot{y}, \dot{z})\), and \((\ddot{x}, \ddot{y})\), respectively.
A pointmap \( O \in \mathbb{R}^{H \times W \times 3} \) is defined as \( H \times W \) points in the camera coordinate system, as shown in Eq.~\ref{eq:pointmap}, where \( \Phi: \mathbb{R}^2 \to \mathbb{R}^3 \) maps \((\ddot{x}, \ddot{y})\) to \((\dot{x}, \dot{y}, \dot{z})\), \( H \) and \( W \) represent the spatial size of the pointmap.
\begin{equation}
O = \{ \Phi(\ddot{x}, \ddot{y}) \mid \ddot{x} = \{0, ..., W-1\}, \ddot{y} = \{0, ..., H-1\} \}
\label{eq:pointmap}
\end{equation}
The DUSt3R model takes a pair of \( H \times W \times 3 \) images from viewpoints \( a \) and \( b \) as inputs and simultaneously regresses the pointmaps \( O^a \) and \( O^b \) in the camera coordinate system of \( a \).
Then, a global alignment strategy is applied to iteratively merge all pointmaps into the same coordinate system and estimate the camera poses for the corresponding images.
\textbf{While it is possible to obtain \(\mathbf{r}_i\) required in Eq.~\ref{eq:ideal_ocl}, optimizing the adversarial example through backpropagation becomes extremely challenging when applied to such an iterative alignment process.}
Therefore, in the following, we will propose an alternative attack strategy that directly enforces orientation consistency based on the pointmaps.

\noindent
\textbf{Enforced projected orientation consistency loss.}
We define the \(i\)-th component of \(\Phi\) by \(\Phi_i\), as is illustrated in Eq.~\ref{eq:pointmap_mappings_i}.
The \(i\)-th channel of the pointmap \( O \), denoted as \( O_i \in \mathbb{R}^{H \times W} \), is then given by Eq.~\ref{eq:pointmap_channel_i}.
\begin{equation}
\dot{x} = \Phi_1(\ddot{x}, \ddot{y}), \, \dot{y} = \Phi_2(\ddot{x}, \ddot{y}), \, \dot{z} = \Phi_3(\ddot{x}, \ddot{y})
\label{eq:pointmap_mappings_i}
\end{equation}
\begin{equation}
O_i = \{ \Phi_i(\ddot{x}, \ddot{y}) \mid \ddot{x} = \{0, ..., W-1\}, \ddot{y} = \{0, ..., H-1\} \}
\label{eq:pointmap_channel_i}
\end{equation}
Note that we only need to consider the part of the pointmap inside the disc region of the kaleidoscopic background.
Taking the \( O_i \) from an arbitrary viewpoint as an example, the pixel coordinates of all points within the disc region in \( O_i \) are defined as a set \( M \).
A line \( l \) inside the disc passing through the center of the disc region divides \( M \) into two parts, \( M_1 \) and \( M_2 \), as illustrated in Fig.~\ref{fig:method}(b).
Define the coordinate variation \(\delta_i(l)\) between \( M_1 \) and \( M_2 \) as Eq.~\ref{eq:coordinate_variations}, where \( |\cdot| \) denotes the number of elements in a set.
Similarly, for a line \( l' \) inside the disc which is perpendicular to \( l \), we can similarly compute its coordinate variation \( \delta_i(l') \). 
\begin{equation}
\delta_i(l) = \frac{\sum_{m_2 \in M_2}{\Phi_i(m_2)}}{| M_2 |}  - \frac{\sum_{m_1 \in M_1}{\Phi_i(m_1)}}{| M_1 |}
\label{eq:coordinate_variations}
\end{equation}
We define the direction of the coordinate flow \(\boldsymbol{\tau}_i \in \mathbb{R}^2\) in Eq.~\ref{eq:flow_direction}, where \( \ddot{\mathbf{u}} \in \mathbb{R}^2\) is the unit normal vector of \( l \) inside the disc pointing from \( M_1 \) to \( M_2 \), and \( \ddot{\mathbf{u}}' \) is the counterpart normal vector of \( l' \).
\begin{equation}
\boldsymbol{\tau}_i = \delta_i(l)\ddot{\mathbf{u}} + \delta_i(l')\ddot{\mathbf{u}}'
\label{eq:flow_direction}
\end{equation}
To account for potential occlusions on the disc, three different lines inside the disc \( l_j  (j=1,2,3)\) are used to estimate coordinate variations from multiple directions, as illustrated in Fig.~\ref{fig:method}(c). 
The angle between \( l_j\) and \( l_{j+1}\) is set to 30 degrees. The average flow direction \( \bar{\boldsymbol{\tau}}_i \) is then computed using Eq.~\ref{eq:average_flow_direction}.
\begin{equation}
\bar{\boldsymbol{\tau}}_i = \frac{1}{3} \sum_{j=1}^{3} \left( \delta_i(l_j)\ddot{\mathbf{u}}_j + \delta_i(l_j')\ddot{\mathbf{u}}'_j \right)
\label{eq:average_flow_direction}
\end{equation}
For two different views \( a \) and \( b \), the projected orientation consistency loss \(\mathcal{L}_{poc}\) is defined as the sum of the cosine similarities between their average flow directions \(\bar{\boldsymbol{\tau}}_i\), as shown in Eq.~\ref{eq:loss}.
\begin{equation}
\mathcal{L}_{poc} = \sum_{i=1}^3 \frac{\bar{\boldsymbol{\tau}}_i^a \cdot \bar{\boldsymbol{\tau}}_i^b}{\|\bar{\boldsymbol{\tau}}_i^a\| \|\bar{\boldsymbol{\tau}}_i^b\|}
\label{eq:loss}
\end{equation}
We visualize the pointmaps \( O_i^a \) and \( O_i^b \)  using heatmaps as shown in Fig.~\ref{fig:method}(d), where brighter areas indicate larger coordinate values in the camera coordinate system.
For a pointmap \( O_i \) of a given viewpoint, it can be observed that the coordinates on the disc increase approximately in a single direction, as indicated by the green and blue arrows in Fig.~\ref{fig:method}(d). In fact, the aforementioned \( \bar{\boldsymbol{\tau}}_i \) can be regarded as a mathematical measure of such a flow direction of the coordinates.  
It is evident that \( \bar{\boldsymbol{\tau}}_i \) is highly correlated with the corresponding camera orientation. 
Intuitively, when the cosine similarity between  \( \bar{\boldsymbol{\tau}}_i^a \) and \( \bar{\boldsymbol{\tau}}_i^b \) are maximized, the camera orientations of views \( a \) and \( b \) tends to be estimated as identical.
Therefore, we maximize the loss function \( \mathcal{L}_{poc} \) in Eq.~\ref{eq:loss} for any two views \( a \) and \( b \) to realize adversarial attacks against the pose estimation.
\textbf{Notably, region partitioning can be directly achieved using several pre-designed masks. }
Moreover, the loss function \( \mathcal{L}_{poc} \) in Eq.~\ref{eq:loss} is simple to implement and can benefit from parallel computation to significantly enhance efficiency.

\noindent
\textbf{Interpretation of $\mathbf{\mathcal{L}_{poc}}$.} 
We hereby interpret the relationship between ${\mathcal{L}_{poc}}$ and the camera orientation vectors.
Following the RDF (right-down-forward) convention~\cite{opencv_library}, we suppose that the kaleidoscopic background lies on the plane \(C: y = 0 \) in the world coordinate system. We denote the projection of the camera orientation vector \(\mathbf{r}_i\) onto plane \(C\) as \(\hat{\mathbf{r}}_i\) in Eq.~\ref{eq:axis_projection}, where \( \mathbf{c} = (0, -1, 0)^\top \) is the normal vector of plane \(C\) and $i=1,2,3$. 
\begin{equation}
\hat{\mathbf{r}}_i = \mathbf{r}_i - (\mathbf{r}_i \cdot \mathbf{c}) \mathbf{c} = 
\begin{bmatrix} 
r_{i1} \hspace{-6pt} & 0 \hspace{-6pt} & r_{i3}
\end{bmatrix}^\top
\label{eq:axis_projection}
\end{equation}
We define the mapping from the world coordinate system to the camera coordinate system as  \(\tilde{\Phi}: \mathbb{R}^3 \to \mathbb{R}^3\)  in Eq.~\ref{eq:mapping_world_to_camera}.
Similar to Eq.~\ref{eq:pointmap_mappings_i}, \(\tilde{\Phi}_i: \mathbb{R}^3 \to \mathbb{R}^1\) is defined as the \(i\)-th component of $\tilde{\Phi}$.
\begin{equation}
\begin{bmatrix}\dot{x} \\ \dot{y} \\ \dot{z}\end{bmatrix} =
\tilde{\Phi}(\begin{bmatrix}x \\ y \\ z\end{bmatrix}) = 
R \begin{bmatrix}x \\ y \\ z\end{bmatrix} + T 
\label{eq:mapping_world_to_camera}
\end{equation}
The Jacobian matrix \( J_{\tilde{\Phi}} \) of the mapping \( \tilde{\Phi} \) can be calculated using Eq.~\ref{eq:mapping_world_to_camera_jacobian}, where \( \nabla \tilde{\Phi}_i \) represents the gradient of the \(i\)-th coordinate in the camera coordinate system with respect to the \(x\), \(y\), and \(z\). Combining Eq.~\ref{eq:R_matrix}  and Eq.~\ref{eq:mapping_world_to_camera_jacobian}, it is evident that \( \nabla \tilde{\Phi}_i = \mathbf{r}_i \).
\begin{equation}
J_{\tilde{\Phi}} = 
R = 
\begin{bmatrix}
r_{11} \hspace{-6pt} & r_{12} \hspace{-6pt} & r_{13} \\ 
r_{21} \hspace{-6pt} & r_{22} \hspace{-6pt} & r_{23} \\ 
r_{31} \hspace{-6pt} & r_{32} \hspace{-6pt} & r_{33}
\end{bmatrix} =
\begin{bmatrix}\nabla \tilde{\Phi}_1 \hspace{-6pt} & \nabla \tilde{\Phi}_2 \hspace{-6pt} & \nabla \tilde{\Phi}_3\end{bmatrix}^\top 
\label{eq:mapping_world_to_camera_jacobian}
\end{equation}
\begin{algorithm}[t!]
\caption{ Kaleidoscopic Background Optimization}
\label{alg:algorithm}
\textbf{Input}: Victim model DUSt3R \(f(\cdot, \cdot)\), differentiable renderer with augmentations \(R(\cdot, \cdot, \cdot, \cdot)\), 3D objects \(O\), environments \(E\), disc object \(o_d\), maximum number of optimization iterations \(T\), color-set clipping frequency \(T_c\) \\
\textbf{Output}: \makebox[0pt][l]{Kaleidoscopic segment image \( I_s \)}
\begin{algorithmic}[1] 
\STATE Initialize \( I_s^0 \) with uniform random noise;
\FOR{$t=0$ to $T$}
\STATE Construct the texture \(I_d\) for \(o_d\) from \(I_s^t\) using Eq.~\ref{eq:seg2disc};
\STATE Randomly select \(o \in O\) and \(e \in E\);
\STATE Render two images \(I_a\) and \(I_b\) from random viewpoints using \(R(o, e, o_d, I_s^0)\) with augmentations;
\STATE Extract Pointmaps \(f(I_a, I_b)\) from DUSt3R;
\STATE Compute orientation consistency loss using Eq.~\ref{eq:loss};
\STATE Update \(I_s^{t+1}\) using Eq.~\ref{eq:optimize};
\IF{$t \bmod T_c = 0$}
\STATE Clip colors in \(I_s^{t+1}\) to the CMYK color space;
\ENDIF
\ENDFOR
\RETURN \( I_s^T \)
\end{algorithmic}
\end{algorithm}
In the world coordinate system, let $L$ be the line within plane $C$, of which the imaging result is line $l$ used in Eq.~\ref{eq:coordinate_variations}. Physically,  $L$ divides the kaleidoscopic background disc evenly into two halves. As a result, the value of \( \delta_i(l) \) computed by Eq.~\ref{eq:coordinate_variations} can be interpreted as the difference  \( \tilde{\Phi}_i(s_2) - \tilde{\Phi}_i(s_1) \), where  \( s_2 \) and \( s_1 \) are the  centroids of the two halves of the disc. 
Suppose \( \mathbf{u} \in \mathbb{R}^3\) is the unit normal vector of  \( L \) inside the disc, the gradient of \( \tilde{\Phi}_i\) along the direction of \( \mathbf{u} \) can be expressed using Eq.~\ref{eq:gradient_u_phi_i}, where \( \| s_1-s_2 \| \) represents the distance between \( s_1 \) and \( s_2 \).
Similarly, we can define line $L'$ and \( \mathbf{u}' \in \mathbb{R}^3\) and calculate the gradient of \( \tilde{\Phi}_i\) along the direction of \( \mathbf{u}' \) as $\nabla_{\mathbf{u}'}$. As such, the gradient of  \( \tilde{\Phi}_i\) in plane $C$ can be expressed by Eq.~\ref{eq:gradient_phi_i}.
\begin{equation}
\nabla_\mathbf{u} = \frac{\tilde{\Phi}_i(s_2) - \tilde{\Phi}_i(s_1)}{\| s_1-s_2 \|} = \frac{\delta_i(l)}{\| s_1-s_2 \|}
\label{eq:gradient_u_phi_i}
\end{equation}
\begin{equation}
\hat{\nabla} \tilde{\Phi}_i = \nabla_\mathbf{u}\mathbf{u} + \nabla_{\mathbf{u}'}\mathbf{u}' = \frac{\delta_i(l){\mathbf{u}} + \delta_i(l'){\mathbf{u}}'}{\| s_1-s_2 \|} \approx \hat{\mathbf{r}}_i 
\label{eq:gradient_phi_i}
\end{equation}
In fact, $\hat{\nabla} \tilde{\Phi}_i $ can be regarded as an estimation of the projection of ${\nabla} \tilde{\Phi}_i $ onto plane $C$ since  vectors \( \mathbf{u} \) and \( \mathbf{u}' \) are defined within the plane, namely $\hat{\mathbf{r}}_i$. 
By assuming orthogonal projection as well as considering the symmetry of the disc, the cosine similarity between $\bar{\boldsymbol{\tau}}_i^a$ and $\bar{\boldsymbol{\tau}}_i^b$ in Eq.~\ref{eq:loss}
can be used as a fair approximation of the cosine similarity between $\hat{\mathbf{r}}_i^a$ and $\hat{\mathbf{r}}_i^b$.
Maximizing  \( \mathcal{L}_{poc} \) is approximately equivalent to maximizing  the sum of cosine similarities of  the projected camera orientation vectors \(\hat{\mathbf{r}}_i\) between different poses, as shown in Eq.~\ref{eq:equal_maximizing}.
\begin{equation}
\arg\max_{I_s} \mathcal{L}_{poc} \Leftrightarrow \arg\max_{I_s} \sum_{i=1}^3 \frac{\hat{\mathbf{r}}_i^a \cdot \hat{\mathbf{r}}_i^b}{\|\hat{\mathbf{r}}_i^a\| \|\hat{\mathbf{r}}_i^b\|}
\label{eq:equal_maximizing}
\end{equation}
\textbf{In fact, $\mathcal{L}_{poc}$ can be regarded as a relaxed version of $\mathcal{L}_{oc}$ by only enforcing camera orientation consistency in the projected plane.} Nevertheless, experiments demonstrate that $\mathcal{L}_{poc}$ achieves a satisfactory attacking effect in practice.


\begin{figure}[t]
  \centering
   \includegraphics[width=1.0\linewidth]{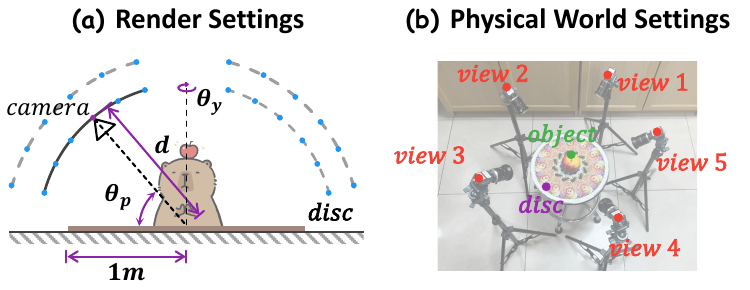}
\caption{(a) The setup for rendering scenes in the digital world. (b) The setup for testing scenes in the physical world.}
   \label{fig:experiment}
\end{figure}

\subsection{The Overall Optimization Process}
\label{sec:overall}
The overall optimization of the kaleidoscopic patterns background is illustrated in Algorithm~\ref{alg:algorithm}. 
We render two images from different viewpoints each time, applying various data augmentation techniques.
For optimization, we maximize the loss computed in Eq.~\ref{eq:loss} and update the kaleidoscopic segment image \( I_s \) as described in Eq.~\ref{eq:optimize}.
Here, \(\alpha = 1/255\) serves as the step size, \(\text{sign}(\cdot)\) indicates the sign function, and \(\text{clip}_{\{0,1\}}(\cdot)\) constrains values within the [0,1] range.
\begin{equation}
I_s^{t+1} = \text{clip}_{\{0,1\}}(I_s^t + \alpha\cdot\text{sign}(\nabla_{I_s} \mathcal{L}_{poc}))
\label{eq:optimize}
\end{equation}
To improve the physical-world feasibility of the kaleidoscopic background, we clip RGB values to the CMYK color space after a specified number of optimization steps.
Refer to the supplementary material for further details.

\begin{figure*}[t!]
  \centering
   \includegraphics[width=\linewidth]{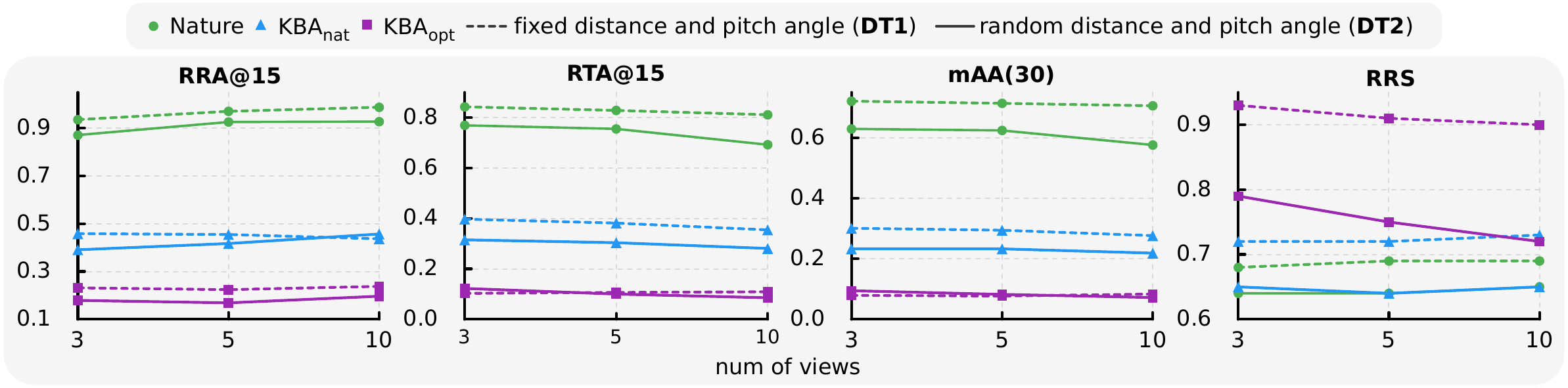}
\caption{
Experimental results of attacking DUSt3R in the digital world with various background discs and view counts. Lower values of \(\text{RRA@}15\), \(\text{RTA@}15\), and \(\text{mAA}(30)\), along with higher RRS values, signify better performance against adversarial attacks.
}
   \label{fig:digital_graph}
\end{figure*}

\begin{table*}[t!]
\centering
\setlength{\tabcolsep}{2.5pt}
\begin{tabular}{c c c c c c c c c c}
\toprule
Methods & Textures & RRA@5 $\downarrow$ & RRA@15 $\downarrow$ & RRA@30 $\downarrow$ & RTA@5 $\downarrow$ & RTA@15 $\downarrow$ & RTA@30 $\downarrow$ & mAA(30) $\downarrow$ & RRS $\uparrow$ \\ 
\midrule
\multirow{3}{*}{DUSt3R} & Nature & 1.00 \scriptsize{±0.02} & 1.00 \scriptsize{±0.00} & 1.00 \scriptsize{±0.00} & 0.98 \scriptsize{±0.03} & 1.00 \scriptsize{±0.00} & 1.00 \scriptsize{±0.00} & 0.95 \scriptsize{±0.01} & 0.62 \scriptsize{±0.01} \\
& KBA\textsubscript{nat} & 0.30 \scriptsize{±0.19} & 0.32 \scriptsize{±0.20} & 0.36 \scriptsize{±0.20} & 0.31 \scriptsize{±0.19} & 0.37 \scriptsize{±0.19} & 0.46 \scriptsize{±0.19} & 0.31 \scriptsize{±0.19} & 0.77 \scriptsize{±0.08} \\
& KBA\textsubscript{opt} & \textbf{0.00} \scriptsize{±0.00} & \textbf{0.00} \scriptsize{±0.00} & \textbf{0.00} \scriptsize{±0.01} & \textbf{0.00} \scriptsize{±0.01} & \textbf{0.02} \scriptsize{±0.03} & \textbf{0.09} \scriptsize{±0.04} & \textbf{0.00} \scriptsize{±0.00} & \textbf{0.94} \scriptsize{±0.02} \\ 
\midrule

\multirow{3}{*}{MASt3R} & Nature & 0.35 \scriptsize{±0.10} & 0.99 \scriptsize{±0.01} & 1.00 \scriptsize{±0.00} & 0.78 \scriptsize{±0.09} & 1.00 \scriptsize{±0.01} & 1.00 \scriptsize{±0.00} & 0.80 \scriptsize{±0.02} & 0.62 \scriptsize{±0.00} \\
& KBA\textsubscript{nat} & 0.14 \scriptsize{±0.10} & 0.45 \scriptsize{±0.19} & 0.51 \scriptsize{±0.20} & 0.32 \scriptsize{±0.16} & 0.51 \scriptsize{±0.19} & 0.63 \scriptsize{±0.20} & 0.36 \scriptsize{±0.15} & 0.72 \scriptsize{±0.08} \\
& KBA\textsubscript{opt} & \textbf{0.02} \scriptsize{±0.04} & \textbf{0.06} \scriptsize{±0.10} & \textbf{0.07} \scriptsize{±0.10} & \textbf{0.05} \scriptsize{±0.08} & \textbf{0.12} \scriptsize{±0.09} & \textbf{0.22} \scriptsize{±0.08} & \textbf{0.05} \scriptsize{±0.08} & \textbf{0.86} \scriptsize{±0.05} \\ 
\midrule

\multirow{3}{*}{Ray Diffusion} & Nature & 0.06 \scriptsize{±0.05} & 0.79 \scriptsize{±0.08} & 0.99 \scriptsize{±0.02} & 0.33 \scriptsize{±0.08} & 0.96 \scriptsize{±0.04} & 0.99 \scriptsize{±0.02} & 0.61 \scriptsize{±0.04} & 0.62 \scriptsize{±0.01} \\
& KBA\textsubscript{nat} & 0.01 \scriptsize{±0.02} & 0.29 \scriptsize{±0.11} & 0.62 \scriptsize{±0.17} & 0.10 \scriptsize{±0.06} & 0.51 \scriptsize{±0.16} & 0.76 \scriptsize{±0.14} & 0.28 \scriptsize{±0.10} & 0.62 \scriptsize{±0.02} \\
& KBA\textsubscript{opt} & \textbf{0.00} \scriptsize{±0.01} & \textbf{0.06} \scriptsize{±0.05} & \textbf{0.21} \scriptsize{±0.10} & \textbf{0.02} \scriptsize{±0.02} & \textbf{0.17} \scriptsize{±0.10} & \textbf{0.39} \scriptsize{±0.12} & \textbf{0.07} \scriptsize{±0.05} & \textbf{0.69} \scriptsize{±0.05} \\ 
\midrule

\multirow{3}{*}{Ray Regression} & Nature & 0.06 \scriptsize{±0.05} & 0.79 \scriptsize{±0.09} & 0.97 \scriptsize{±0.04} & 0.26 \scriptsize{±0.10} & 0.90 \scriptsize{±0.07} & 0.97 \scriptsize{±0.04} & 0.60 \scriptsize{±0.05} & 0.62 \scriptsize{±0.01} \\
& KBA\textsubscript{nat} & 0.01 \scriptsize{±0.02} & 0.27 \scriptsize{±0.11} & 0.59 \scriptsize{±0.16} & 0.08 \scriptsize{±0.04} & 0.49 \scriptsize{±0.15} & 0.70 \scriptsize{±0.13} & 0.26 \scriptsize{±0.09} & 0.62 \scriptsize{±0.02} \\
& KBA\textsubscript{opt} & \textbf{0.00} \scriptsize{±0.00} & \textbf{0.10} \scriptsize{±0.07} & \textbf{0.35} \scriptsize{±0.16} & \textbf{0.03} \scriptsize{±0.02} & \textbf{0.26} \scriptsize{±0.12} & \textbf{0.49} \scriptsize{±0.14} & \textbf{0.11} \scriptsize{±0.06} & \textbf{0.66} \scriptsize{±0.06} \\ 
\midrule

\multirow{3}{*}{RelPose++} & Nature & 0.33 \scriptsize{±0.11} & 0.57 \scriptsize{±0.10} & 0.57 \scriptsize{±0.10} & 0.25 \scriptsize{±0.09} & 0.52 \scriptsize{±0.10} & 0.57 \scriptsize{±0.10} & 0.43 \scriptsize{±0.08} & 0.62 \scriptsize{±0.01} \\
& KBA\textsubscript{nat} & 0.06 \scriptsize{±0.04} & 0.30 \scriptsize{±0.10} & 0.46 \scriptsize{±0.12} & 0.10 \scriptsize{±0.05} & 0.40 \scriptsize{±0.13} & 0.55 \scriptsize{±0.11} & 0.25 \scriptsize{±0.08} & 0.61 \scriptsize{±0.01} \\
& KBA\textsubscript{opt} & \textbf{0.04} \scriptsize{±0.04} & \textbf{0.20} \scriptsize{±0.08} & \textbf{0.30} \scriptsize{±0.09} & \textbf{0.08} \scriptsize{±0.04} & \textbf{0.27} \scriptsize{±0.10} & \textbf{0.39} \scriptsize{±0.10} & \textbf{0.17} \scriptsize{±0.07} & \textbf{0.66} \scriptsize{±0.05} \\ 
\midrule

\multirow{3}{*}{PoseDiffusion} & Nature & 0.39 \scriptsize{±0.11} & 0.64 \scriptsize{±0.10} & 0.65 \scriptsize{±0.10} & 0.41 \scriptsize{±0.11} & 0.61 \scriptsize{±0.11} & 0.64 \scriptsize{±0.10} & 0.51 \scriptsize{±0.09} & 0.62 \scriptsize{±0.00} \\
& KBA\textsubscript{nat} & \textbf{0.02} \scriptsize{±0.02} & 0.19 \scriptsize{±0.06} & 0.40 \scriptsize{±0.10} & 0.10 \scriptsize{±0.04} & 0.42 \scriptsize{±0.10} & 0.61 \scriptsize{±0.11} & 0.19 \scriptsize{±0.05} & 0.62 \scriptsize{±0.03} \\
& KBA\textsubscript{opt} & \textbf{0.02} \scriptsize{±0.02} & \textbf{0.14} \scriptsize{±0.06} & \textbf{0.26} \scriptsize{±0.12} & \textbf{0.04} \scriptsize{±0.03} & \textbf{0.23} \scriptsize{±0.11} & \textbf{0.40} \scriptsize{±0.12} & \textbf{0.11} \scriptsize{±0.06} & \textbf{0.65} \scriptsize{±0.05} \\ 
\bottomrule
\end{tabular}
\caption{
Experimental results of attacking various camera pose estimation models with different background discs in the physical world. 
Each cell contains two values: the larger value represents the mean of the metric across all samples, while the smaller value indicates the standard deviation of the metric across different object categories. 
Bold values indicate the best performance of adversarial attacks.}
\label{tab:physical}
\end{table*}
\section{Experiments}
\label{sec:experiments}
To validate the effectiveness of our approach, we perform a series of experiments in both the digital and physical worlds. 
In the digital world, we focus on optimizing adversarial backgrounds and conducting ablation studies. 
In contrast, the physical world experiments primarily evaluate the effectiveness and generalizability of our approach in attacking various camera pose estimation models under complex real-world scenarios. 
For these experiments, \textbf{Nature} refers to a desktop texture, with its radially symmetric version denoted as \textbf{KBA\textsubscript{nat}} and its optimized counterpart as \textbf{KBA\textsubscript{opt}}. 
We use consistent evaluation metrics across both types of experiments for a comprehensive comparison.

\noindent
\textbf{Evaluation metircs.}
Following ~\cite{Wang2024dust3r, vincent2024mast3r, zhang2024raydiffusion, Wang2023PoseDiffusion, lin2024relposepp}, we evaluated the accuracy of pose estimation using Relative Rotation Accuracy (RRA), Relative Translation Accuracy (RTA), and mean Average Accuracy (mAA).
Specifically, RRA compares the relative rotation \(R_i R_j^\top\) from the \(i\)-th to the \(j\)-th camera with the ground truth \(R_i^\star R_j^{\star\top}\), while RTA measures the angle between the predicted vector \(T_{ij}\) and the ground truth vector \(T_{ij}^\star\) pointing from camera \(i\) to camera \(j\).
We report \( \text{RTA@}\gamma \) and \( \text{RRA@}\gamma \) (\(\gamma \in \{5, 15, 30\}\)), representing the percentage of camera pairs with RRA or RTA values below the threshold \(\gamma\).
Furthermore, we compute the \(\text{mAA}(30)\), which is defined as the area under the accuracy curve of angular differences at \( min(\text{RRA@}30, \text{RTA@}30) \).
Beyond these three standard metrics, we introduce a custom Relative Rotation Similarity (RRS) metric, leveraging cosine similarity to assess the similarity between different predicted relative rotations \(R_i R_j^\top\). 
An RRS value close to \(1\) signifies high consistency in pose orientations.

\begin{figure*}[t!]
	\centering
	\includegraphics[width=\linewidth]{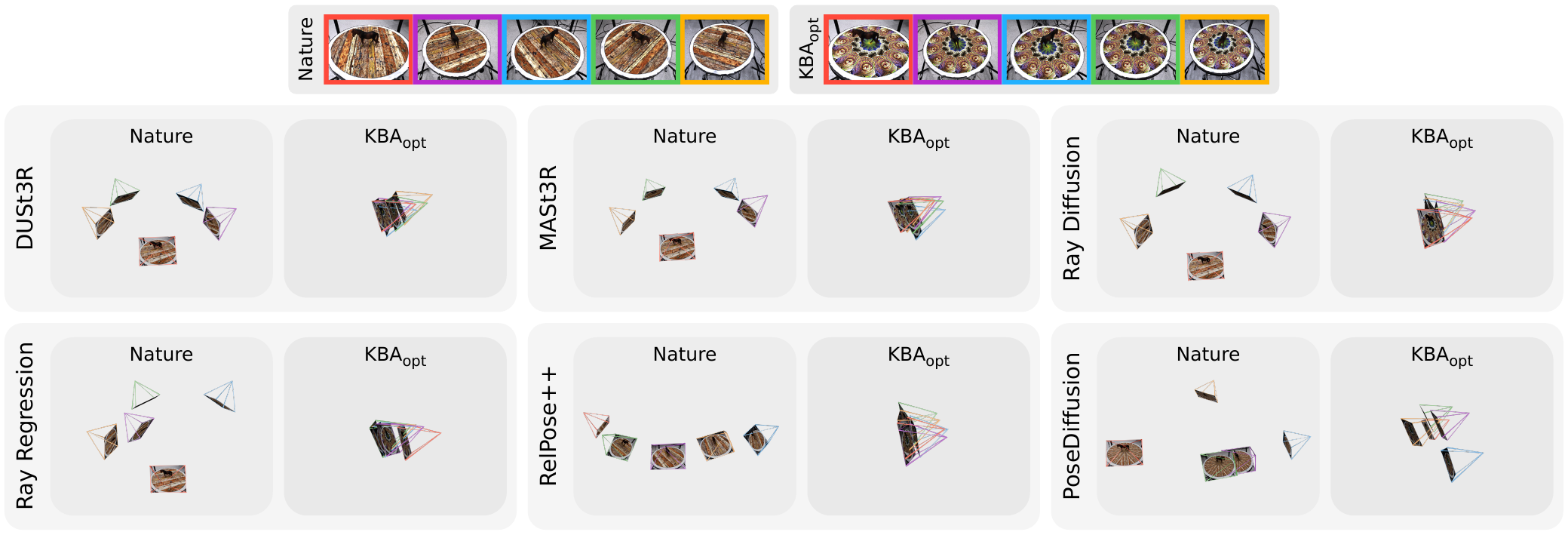}
	\caption{
		Visualization of experimental results for discs with varying backgrounds across different camera pose estimation models in the physical world. 
		The color of the image borders corresponds to the color of the associated pose pyramid.
	}
	\label{fig:vis_physical}
\end{figure*}

\subsection{Experiments in the Digital World}
In this section, we first introduce the data used for adversarial attacks, followed by the test setup in the digital world. 
Finally, we present the camera pose estimation results under different backgrounds for 3, 5, and 10 views.

\noindent
\textbf{Attack data and parameters.}
We use six HDRI images of indoor and outdoor scenes from Polyhaven~\cite{polyhaven}, each mapped onto a spherical mesh to create realistic backgrounds. 
Additionally, we select 32 3D objects from 20 categories in OmniObject3D~\cite{wu2023omniobject3d} for adversarial attacks. 
Both KBA\textsubscript{nat} and KBA\textsubscript{opt} are constructed with \( N = 12 \) segments. 
At each optimization step, objects and backgrounds are randomly selected to construct the scene. 

\noindent
\textbf{Test setups.}
For the test, we additionally select \(10\) HDRI images from Polyhaven and \(25\) objects spanning \(25\) categories in the OmniObject3D dataset, none of which are included in the attack data.
The rendered scene, depicted in Fig.~\ref{fig:experiment}(a), includes a 3D object, background disc, and environment, with the disc radius set to \(1\) m, objects centered and scaled within a \(0.8 \times 0.8 \times 0.8\) m bounding box, and the rendering camera positioned at distance \(d\) facing the disc center, with pitch angle \(\theta_p\) and yaw angle \(\theta_y\) defining its orientation.
In order to increase the diversity of experiments and the generalization of results, we configured \(6 \times 36 \times 6 = 1296\) parameter combinations using six pitch angles (ranging from \(10^\circ\) to \(85^\circ\)), \(36\) yaw angles (in increments of \(10^\circ\)), and six distances (ranging from \(2.0\) m to \(3.0\) m).
Images and masks for the 3D objects, discs, and environments were rendered and combined during testing to produce the final image.
We design two testing scenarios, \textbf{DT1} and \textbf{DT2}, to assess the impact of different background discs on camera pose estimation. 
In \textbf{DT1}, the pitch angle is fixed at \(55^\circ\) and the distance at \(2.4\) m, while varying the yaw angles. 
In \textbf{DT2}, \(1296\) camera poses are randomly selected for a more comprehensive evaluation. 
Each scenario includes four samples per object-environment combination, resulting in a total of \(25 \times 10 \times 4 = 1000\) samples.

\noindent
\textbf{Experimental results.}
The results in Fig.~\ref{fig:digital_graph} demonstrate that KBA\textsubscript{nat} is already capable of significantly reducing the RRA, RTA, and mAA metrics compared to the natural background.
KBA\textsubscript{opt} obtains radially symmetric textures through optimization, showing a marked improvement in attack effectiveness compared to KBA\textsubscript{nat}.
It is noteworthy that, although our adversarial attack optimization process is conducted on a pair of images from two views, stable adversarial attack effectiveness is observed when using 3, 5, or 10 images for camera pose estimation, regardless of whether in the same longitude DT1 or the more random DT2 settings.
Regarding the RRS metric, both KBA\textsubscript{nat} and KBA\textsubscript{opt} enhance camera orientation similarity in DT1 with yaw angle changes, with KBA\textsubscript{opt} achieving a similarity close to $0.9$.
Although enforcing identical camera orientations across all views becomes challenging as the number of images increases in the DT2 setting, our method still significantly disrupts camera pose estimation, as indicated by the RRA, RTA, and mAA metrics.
We also conduct adversarial transferability experiments in the digital world. 
The results show that the radially symmetric textures optimized by KBA\textsubscript{opt} exhibit strong transferability across various models. 
The details are provided in the supplementary material.

\begin{figure*}[ht]
	\centering
	\includegraphics[width=\linewidth]{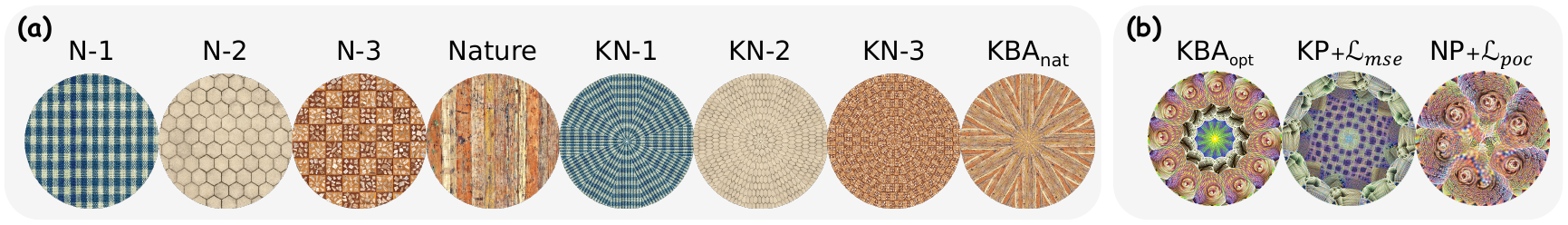}
	\caption{
		(a) Naturally symmetrical textures found in nature (denoted as N-1,2,3) and unoptimized radially symmetrical textures derived from them (denoted as KN-1,2,3). 
		(b) Textures optimized using different combinations of loss functions and background patterns.
	}
	\label{fig:experiment_discuss}
\end{figure*}

\subsection{Experiments in the Physical World}
In this section, we first describe the physical-world test setup, followed by a detailed analysis of the results obtained with various background textures across multiple models. Additionally, we present visualizations of the camera pose estimation outcomes based on a set of captured images, further illustrating the effectiveness of our approach.

\noindent
\textbf{Test setups.}
We selected 24 different 3D objects, including vegetables, fruits, animals, and vehicles.
To ensure complete capture of the disc while maintaining good imaging quality for the objects, we crafted two discs: one with a radius of \(15\) cm for objects ranging from \(10\) to \(20\) cm, and another with a radius of \(20\) cm for objects ranging from \(20\) to \(30\) cm.
As shown in Fig.~\ref{fig:experiment}(b), five industrial cameras are evenly distributed around the disc, with their lenses directed toward the center at distances ranging from \(20\) to \(50\) cm.
Under normal indoor lighting conditions, we calibrated the cameras using a calibration board and simultaneously captured object-centric images from five different viewpoints.
We captured five groups of images for each object, adjusting the camera poses between groups to ensure data diversity.

\begin{table}[t]
	\centering
	\setlength{\tabcolsep}{1.8pt}
	\begin{tabular}{c c c c c}
		\toprule
		Textures & RRA@15 $\downarrow$ & RTA@15 $\downarrow$ & mAA(30) $\downarrow$ & RRS $\uparrow$ \\ 
		\midrule
		N-1 & 0.89 \scriptsize{±0.02} & 0.78 \scriptsize{±0.05} & 0.63 \scriptsize{±0.02} & 0.64 \scriptsize{±0.01} \\
		N-2 & 0.91 \scriptsize{±0.02} & 0.78 \scriptsize{±0.07} & 0.66 \scriptsize{±0.05} & 0.64 \scriptsize{±0.01} \\
		N-3 & 0.97 \scriptsize{±0.01} & 0.87 \scriptsize{±0.06} & 0.78 \scriptsize{±0.05} & 0.64 \scriptsize{±0.01} \\
		Nature & 0.97 \scriptsize{±0.01} & 0.83 \scriptsize{±0.07} & 0.71 \scriptsize{±0.05} & 0.64 \scriptsize{±0.01} \\
		\midrule
		KN-1 & 0.70 \scriptsize{±0.04} & 0.58 \scriptsize{±0.05} & 0.45 \scriptsize{±0.02} & 0.64 \scriptsize{±0.01} \\
		KN-2 & 0.56 \scriptsize{±0.05} & 0.47 \scriptsize{±0.04} & 0.38 \scriptsize{±0.03} & 0.66 \scriptsize{±0.01} \\
		KN-3 & 0.45 \scriptsize{±0.04} & 0.40 \scriptsize{±0.03} & 0.32 \scriptsize{±0.02} & 0.67 \scriptsize{±0.01} \\
		KBA\textsubscript{nat} & 0.46 \scriptsize{±0.06} & 0.38 \scriptsize{±0.03} & 0.29 \scriptsize{±0.02} & 0.67 \scriptsize{±0.01} \\
		\midrule
		KBA\textsubscript{opt} & \textbf{0.22} \scriptsize{±0.07} & \textbf{0.11} \scriptsize{±0.02} & \textbf{0.08} \scriptsize{±0.02} & \textbf{0.86} \scriptsize{±0.01} \\
		KP + $\mathcal{L}_\text{mse}$ & 0.27 \scriptsize{±0.12} & 0.16 \scriptsize{±0.04} & 0.13 \scriptsize{±0.04} & 0.76 \scriptsize{±0.03} \\
		NP + $\mathcal{L}_\text{poc}$ & 0.59 \scriptsize{±0.08} & 0.35 \scriptsize{±0.02} & 0.29 \scriptsize{±0.02} & 0.70 \scriptsize{±0.01} \\
		\bottomrule
	\end{tabular}
	\caption{
		Camera pose estimation results on DUSt3R with different backgrounds using the DT1 5 images setting.
		Bold values indicate the best performance among adversarial attacks.
	}
	\label{tab:discussion}
\end{table} 

\noindent
\textbf{Experimental results.}
We evaluated the pose estimation performance of Nature, KBA\textsubscript{nat}, and KBA\textsubscript{opt} in the physical world, across both the white-box model DUSt3R~\cite{Wang2024dust3r} and the black-box models MASt3R~\cite{vincent2024mast3r}, RayDiffusion~\cite{zhang2024raydiffusion}, RayRegression~\cite{zhang2024raydiffusion}, PoseDiffusion~\cite{Wang2023PoseDiffusion}, and RelPose++~\cite{lin2024relposepp}, as shown in Tab.~\ref{tab:physical}.
The experimental results show that KBA\textsubscript{nat} achieves notable attack effectiveness across various pose estimation models. 
In comparison, the optimized KBA\textsubscript{opt} exhibits significantly higher performance with smaller standard deviations across metrics, indicating more consistent adversarial attack effectiveness across various object categories.
We visualize the camera pose estimation results of various methods on five images with differing perspectives, where the disks correspond to Nature and KBA\textsubscript{opt}, as shown in Fig.~\ref{fig:vis_physical}. 
KBA\textsubscript{opt} exhibits stable adversarial attack performance and effective transferability across multiple black-box models. 
Notably, KBA\textsubscript{opt} maximizes the consistency of pose orientation during the attack without explicitly constraining camera positions.
However, physical experiments reveal that KBA\textsubscript{opt} leads to nearly overlapping camera positions, approximating the situation where five images are captured from a single location, further demonstrating the effectiveness of our adversarial attacks.
We further visualize the camera pose estimation results in more complex scenarios, including cases where various objects are placed on a disc, the disc is off-centered in the image, and scene-centric environments, to demonstrate the generalizability of our approach. 
These visualizations, along with a discussion on the performance of existing patch defense methods against our attacks, are provided in the supplementary material.

\subsection{Additional Ablation Studies}
In this section, we validate the effectiveness of the radially symmetric texture pattern and the proposed loss function through ablation studies.
We first evaluate several symmetric textures found in nature, including woven fabric, hexagonal tile patterns, and quadrilateral tile patterns, as depicted in Fig.~\ref{fig:experiment_discuss} (a).
The results on DUSt3R in Tab.~\ref{tab:discussion} indicate that such natural symmetry has limited impact on camera pose estimation.
Building on these natural textures, we construct the corresponding kaleidoscopic backgrounds, as illustrated by KN-1, KN-2, and KN-3 in Fig.~\ref{fig:experiment_discuss} (a).
Experiments demonstrate that radially symmetric textures constructed from various natural patterns have already shown significant interference in camera pose estimation.
To validate the effectiveness of KBA\textsubscript{opt} (KP + $\mathcal{L}_\text{poc}$), we further test a typical patch-based optimization applied to the entire image (NP) and a simple loss strategy minimizing the output MSE across views (\(\mathcal{L}_{mse}\)).
The textures obtained from the optimization of different method combinations are shown in Fig.~\ref{fig:experiment_discuss} (b).
Tab.~\ref{tab:discussion} indicates that combining NP and \(\mathcal{L}_{mse}\) with our \(\mathcal{L}_\text{poc}\) and KP reduces the effectiveness of adversarial attacks, further validating the roles of radially symmetric textures and our loss function.


\section{Conclusions}
\label{sec:conclusion}
In this paper, we propose a method for constructing the multi-fold radial symmetric adversarial kaleidoscopic background that exhibits notable similarity across multiple viewpoints to attack camera pose estimation models. We propose a projected orientation consistency loss for optimizing the kaleidoscopic background based on pointmaps, leading to further improvements in attack effectiveness. 
Experimental results demonstrate that our approach effectively attacks camera pose estimation models under both white-box and black-box settings in the digital and physical worlds, while maintaining strong robustness across varying scenes and camera configurations.

\section{Acknowledgements}
\label{sec:acknowledgements}
This work was supported by the National Natural Science Foundation of China (62376024), the National Science and Technology Major Project (2022ZD0117902), and the Fundamental Research Funds for the Central Universities (FRF-TP-22-043A1).

{
    \small
    \bibliographystyle{ieeenat_fullname}
    \bibliography{main}
}

\end{document}